\documentclass[runningheads]{llncs}

\usepackage[T1]{fontenc}
\usepackage[utf8]{inputenc}

\usepackage{graphicx}
\usepackage{amsmath}
\usepackage{amssymb}
\usepackage{color}
\usepackage[section]{placeins}
\usepackage{booktabs}
\usepackage{url}
\makeatletter
\def\ps@headings{\let\@mkboth\@gobbletwo
  \let\@oddfoot\@empty\let\@evenfoot\@empty
  \def\@evenhead{\normalfont\small\hspace{\headlineindent}%
                 \leftmark\hfil}
  \def\@oddhead{\normalfont\small\hfil\rightmark\hspace{\headlineindent}}
  \def\chaptermark##1{}%
  \def\sectionmark##1{}%
  \def\subsectionmark##1{}}
\makeatother
\begin{document}

\title{Enhancing Explainable Cardiac Diagnosis with Guide-Grounded Multimodal LLMs}
\titlerunning{Enhancing XAI using Guide-Grounded Multimodal LLMs}

\author{Hai-Nam Duy Vuong\inst{1,4} \and Duy-Anh Bui\inst{1} \and
Trong-Nghia Nguyen\inst{1} \and \\ Kim-Ngan Thi Nguyen\inst{1}  \and 
Trang Mai Xuan\inst{2}\thanks{Corresponding author.} \and Tien-Cuong Nguyen\inst{3}  \and \\ Van-Dem Pham\inst{5} \and Thien Van Luong\inst{1}}

\institute{Business AI Lab, College of Technology, National Economics University, Vietnam\\
\and
A2I Lab, Phenikaa School of Computing, Phenikaa University, Hanoi, Vietnam\\
\and
VNPT AI, VNPT Group, Hanoi, Vietnam\\
\and
FPT University, Hanoi, Vietnam\\
\and
Department of Pediatrics, Hospital of University Medicine and Pharmacy, Vietnam National University Hanoi, Vietnam\\
\email{namvdh.bai@st.neu.edu.vn, anhbd.bai@st.neu.edu.vn, nghiant@neu.edu.vn, ngannguyen@neu.edu.vn, trang.maixuan@phenikaa-uni.edu.vn, nguyentiencuong@vnpt.vn, dempv.ump@vnu.edu.vn, thienlv@neu.edu.vn}
}

\authorrunning{V.\ D.\ H.\ Nam \textit{et al.}}

\maketitle

\begin{abstract}

The electrocardiogram (ECG) is a cornerstone of cardiac assessment, yet clinical deployment of deep learning models remains constrained by limited interpretability and the hallucination risk of large language models (LLMs). Existing CNN+Grad-CAM+multimodal LLM frameworks can generate ECG reports, but their explanations are often only weakly grounded in established diagnostic criteria, reducing trustworthiness and reproducibility. We propose a guide-grounded multimodal framework that explicitly anchors report generation in curated clinical knowledge. A convolutional neural network (CNN) and Grad-CAM first produce class probabilities and class-specific heatmaps from 12-lead ECG images. In parallel, authoritative ECG textbooks and guideline materials are distilled offline into a structured ECG Interpretation Guide, which is injected as a fixed knowledge block for every sample. Conditioned on the ECG image, Grad-CAM overlay, CNN-derived fact pack, and the injected guide, a multimodal LLM generates structured diagnostic reports with guideline-consistent terminology and criteria usage. Experiments on the full PTB-XL test set demonstrate that guide grounding improves semantic quality and perceived consistency of generated reports while preserving competitive classification performance. In particular, our method increases the average BERTScore of generated impressions from $0.818$ to $0.953$ relative to a strong CNN+Grad-CAM+MLLM baseline, indicating closer alignment with reference reports. These findings suggest that injecting a distilled interpretation guide into the multimodal prompting pipeline offers a practical pathway to reduce hallucinations and enhance the clinical plausibility of LLM-based ECG explanations, bringing explainable cardiac diagnosis closer to real-world deployment.

\end{abstract}

\section{Introduction}

In the past few years, large language models (LLMs) have begun to support clinicians in documenting and interpreting medical data. In ECG analysis, multimodal LLMs can convert model outputs and visual evidence into natural-language diagnostic reports, potentially reducing clinician workload. Nevertheless, hallucinated or weakly grounded explanations remain a major barrier to clinical adoption, particularly in high-stakes decision-making. 

Electrocardiography is a cornerstone in cardiology for assessing cardiac function and diagnosing arrhythmias, myocardial infarction, conduction blocks and other heart diseases. Large-scale ECG databases such as PTB-XL~\cite{wagner2020ptbxl,strodthoff2021deeplearningptbxl} have enabled the development of powerful deep learning models that achieve cardiologist-level performance on several diagnostic tasks~\cite{rajpurkar2017cardiologist}. Despite these advances, clinical adoption remains limited due to the lack of transparency of deep neural networks and the risk of erroneous or hallucinated explanations produced by large language models (LLMs). In order to resolve this drawback, we have used a method called Explainable AI (XAI). Explainable AI (XAI) techniques such as Gradient-weighted Class Activation Mapping (Grad-CAM)~\cite{selvaraju2017gradcam} provide visual insight into the regions of an ECG that contribute most to a model's prediction, and have been successfully applied to ECG analysis~\cite{hicks2021ecggradcam,jahmunah2022mi_gradcam}. However, such visualizations still require substantial domain expertise to interpret, and they do not directly translate into textual rationales that can be integrated into clinical reports.

In recent years, deep learning has been widely applied to ECG analysis, achieving state-of-the-art performance in arrhythmia classification and myocardial infarction detection~\cite{rajpurkar2017cardiologist,strodthoff2021deeplearningptbxl}. Convolutional neural networks (CNNs) and residual architectures are particularly effective in capturing morphological features from 1D signals or 2D ECG images, and large-scale datasets such as PTB-XL~\cite{wagner2020ptbxl,strodthoff2021deeplearningptbxl} provide over 20\,000 10-second 12-lead ECGs with rich diagnostic labels to support benchmarking of such models. Despite strong predictive accuracy, these networks often operate as black boxes, which is problematic in high-stakes clinical settings where accountability and interpretability are essential; thus, explainable AI (XAI) methods such as Grad-CAM~\cite{selvaraju2017gradcam}, LIME~\cite{ribeiro2016lime} and SHAP~\cite{lundberg2017shap} have been adapted to ECG models~\cite{hicks2021ecggradcam,jahmunah2022mi_gradcam}, with Grad-CAM producing class-specific heatmaps that can align with clinically meaningful intervals. Building on this, recent work has begun exploring multimodal LLMs (MLLMs) to generate natural language explanations and diagnostic reports from ECG images~\cite{wu2024ecgllm,zhao2024ecgchat,liu2024teachmlmecg}, typically combining a CNN classifier, Grad-CAM heatmaps, and an MLLM that produces free-text reports given the ECG image and model outputs. However, these systems largely rely on the internal knowledge of the LLM and are often only weakly grounded in formal ECG textbooks \cite{hampton2019ecg,sajjan2013learn} or clinical guidelines, limiting trustworthiness and reproducibility; moreover, general-purpose LLMs such as GPT-4~\cite{openai2023gpt4} and Gemini are known to hallucinate plausible-sounding but factually incorrect medical statements. To mitigate this issue, we propose a guide-grounded method: we distill authoritative ECG literature into a structured \emph{ECG Interpretation Guide} and inject it as a fixed knowledge block for \emph{every} sample, encouraging guideline-consistent phrasing and criteria usage while remaining visually anchored to Grad-CAM evidence. In parallel, LLM-based automated paradigms that use strong LLMs such as Gemini to score or compare model outputs have been shown to correlate with human preferences in language generation tasks, providing an additional automatic evaluation signal for explanation quality~\cite{zheng2024judging}.

In this paper, we propose a guide-grounded multimodal framework that explicitly grounds MLLM-generated ECG reports in curated textbooks \cite{hampton2019ecg,sajjan2013learn} and clinical guidelines. Our contributions are threefold:
\begin{enumerate}
  \item We introduce a three-stage pipeline that integrates CNN-based ECG classification, Grad-CAM visual explanations, and guide-grounded multimodal report generation. The final reports are explicitly conditioned on a distilled ECG Interpretation Guide injected into the prompt for every sample.
  \item We construct an ECG Interpretation Guide from curated medical textbooks \cite{hampton2019ecg,sajjan2013learn} and guidelines via an offline distillation procedure, designed to preserve clinically relevant content while removing noise and redundancy.
  \item We conduct a systematic comparison between our guide-grounded framework and a strong baseline that uses CNN, Grad-CAM and MLLM without the guide, on both a 200-sample subset and the full PTB-XL test set. Using standard classification metrics, BERTScore~\cite{zhang2020bertscore}, and an LLM-based Automated evaluation of report quality, we show that guide grounding improves semantic quality and perceived consistency while maintaining competitive classification performance.
\end{enumerate}

\section{Proposed Method}
\label{sec:method}
\begin{figure}[t]
  \centering
  \includegraphics[width=\textwidth]{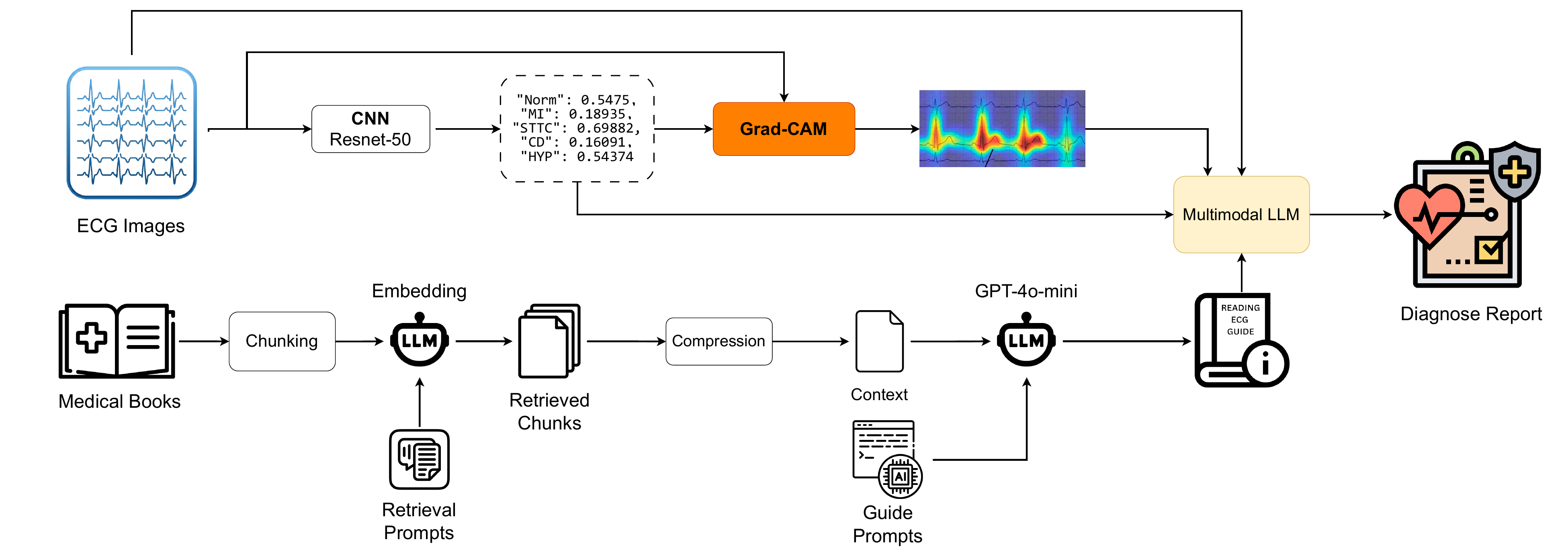}
  \caption{Guide-grounded multimodal report generation.}
  \label{fig:finalstage}
\end{figure}

We propose a three-stage pipeline for explainable ECG diagnosis that combines CNN-based prediction, Grad-CAM visual evidence, and guide-grounded multimodal report generation. The key idea is to keep the diagnosis evidence visually anchored (ECG + Grad-CAM) while improving textual explainability by injecting a distilled ECG Interpretation Guide into the multimodal prompt for every sample to reduce hallucinations.

\subsection{ECG Preprocessing, CNN Diagnosis and Grad-CAM Evidence}
\label{subsec:method_stage1}

\textbf{Signal preprocessing and image rendering.}
Given a 12-lead ECG waveform, we apply light denoising to reduce common artifacts: moving-average smoothing, a notch filter to suppress power-line interference, and a high-pass Butterworth filter to mitigate baseline wander~\cite{butterworth1930theory}. The processed waveform is then rendered into a single ECG image by plotting the 12 leads as stacked traces with grid lines, producing the visual input used in subsequent stages.

\textbf{CNN classifier.}
Let $x$ denote the rendered ECG image. A CNN classifier $f_{\theta}$ outputs diagnostic probabilities
\begin{equation}
\mathbf{p} = f_{\theta}(x) \in [0,1]^K,
\end{equation}
where $K{=}5$ corresponds to the PTB-XL diagnostic superclasses. Since each ECG may contain multiple superclasses, we use a multi-label setup with sigmoid outputs. We employ a ResNet-50 backbone with a task-specific classification head.

\textbf{Grad-CAM visual evidence.}
To expose which waveform regions support the CNN decision, we apply Grad-CAM~\cite{selvaraju2017gradcam} on the last convolutional layer. For the top 3 predicted class, Grad-CAM forms a class-specific heatmap by weighting activation maps with spatially averaged gradients of the class score, followed by ReLU and normalization. The heatmap is upsampled to the original ECG image resolution and overlayed to produce an interpretable visual explanation used as primary evidence in Stage~3.

\subsection{ECG Interpretation Guide Distillation}
\label{subsec:method_stage2}

To ground generated explanations in authoritative medical knowledge, we distill curated ECG textbooks \cite{hampton2019ecg,sajjan2013learn} and guideline materials into a single structured \emph{ECG Interpretation Guide}. The guide is created offline and later injected into the multimodal prompt in Stage~3 for every sample.

\textbf{Chunking.}
We extract page-level text from each medical book and concatenate pages into a raw corpus. To accommodate long-context processing, we split the corpus into large chunks (up to 700{,}000 characters per chunk; three chunks in our setup), which reduces fragmentation while remaining within modern LLM context limits.

\textbf{Embedding and retrieval.}
We embed the page-level chunks using an LLM-based embedder and store the resulting vectors in a database. We then use a set of \emph{retrieval prompts} that instruct LLM (gpt-5-mini) to identify and request the most relevant pages/sections from the indexed corpus for guide construction. These prompts specify the required evidence scope, so the retriever returns the corresponding \emph{retrieved chunks} (pages/paragraphs) to be included as supporting material for the guide.

\textbf{Compression.}
The \emph{retrieved chunks} are then compressed to remove duplicated lines, repeated headers/footers, and obvious boilerplate. The instruction explicitly forbids summarization and forbids removing medically relevant content; the goal is only to reduce noise while preserving clinical meaning. The resulting compressed context is concatenated into a single clean evidence context.

\begin{figure}[t]
  \centering
  \includegraphics[width=\textwidth]{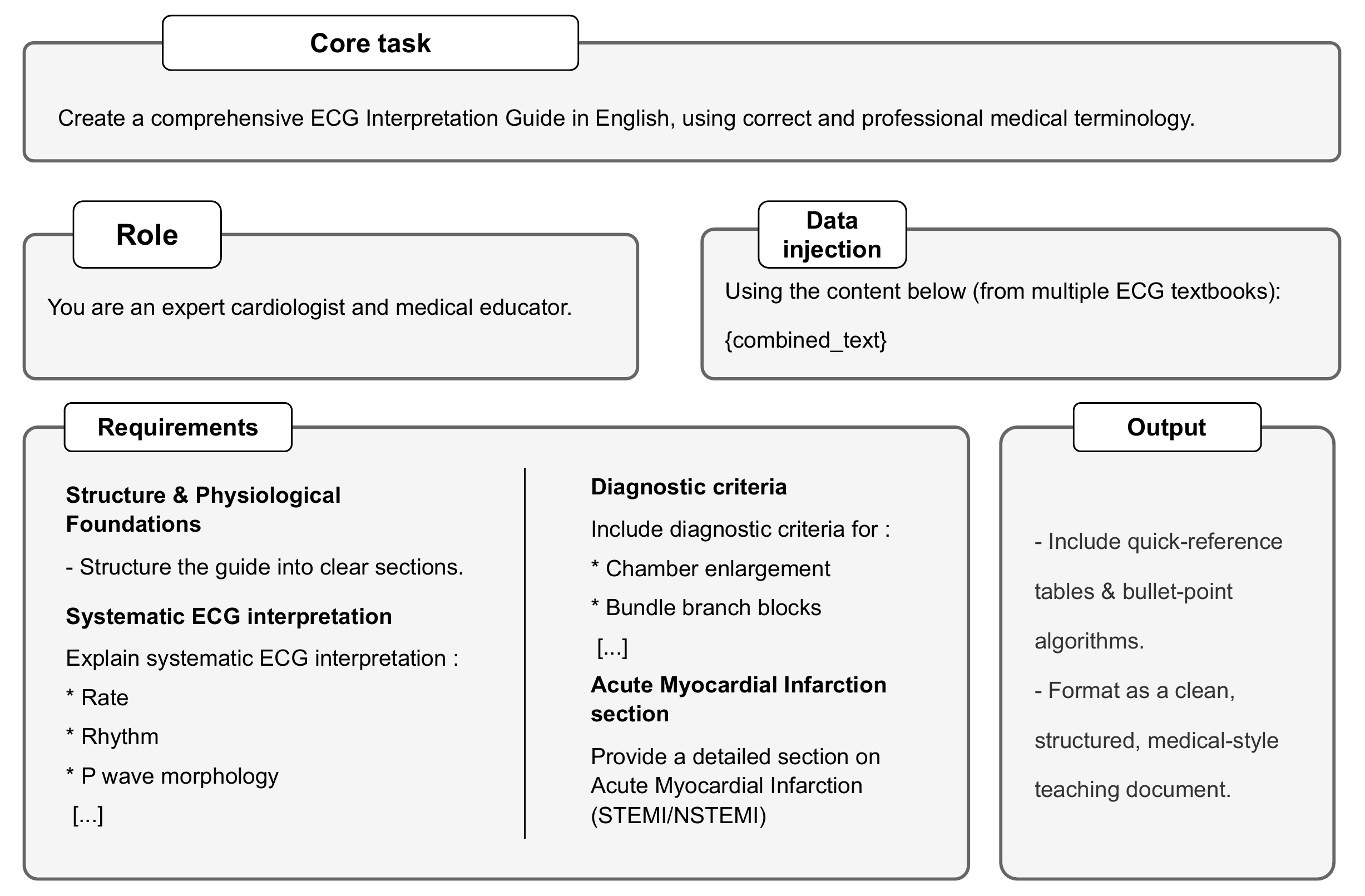}
  \caption{Prompt structure for ECG guide.}
  \label{fig:prompting} 
\end{figure}

\textbf{Guide synthesis.}
A final LLM call (gpt-4o-mini) combines the compressed retrieved context and the fixed \emph{guide prompts} (Fig.~\ref{fig:prompting}) to synthesize a clean, sectioned teaching document in professional medical terminology. This offline synthesis produces a fixed \emph{ECG Interpretation Guide} that is reused as a consistent grounding reference for all subsequent diagnostic queries.

\textbf{Guide prompts.}
Lightly compress redundant lines while preserving all medical meaning. Do not summarize. Do not remove clinically relevant content. Create a comprehensive ECG Interpretation Guide in English using professional medical terminology. Structure into clear sections; include ECG physiology, systematic interpretation steps, diagnostic criteria for major pathologies, and a detailed STEMI/NSTEMI section; include quick-reference tables and algorithms. Output only the final guide.

We provide a public ECG Interpretation Guide created in Stage 2 (this work). The guide is synthesized offline from a curated set of ECG references (listed in References) and released in our code repository with versioning.\footnote{\url{https://github.com/NonomiyaIzumi/XAI_ECG}.} The guide is an original, structured summary intended for prompting and is not a verbatim reproduction of any single source.

\subsection{Guide-Grounded Multimodal Report Generation}
\label{subsec:method_stage3}

In the final stage, a multimodal LLM generates a structured diagnostic report conditioned on four inputs:
(i) the processed ECG image, (ii) the Grad-CAM overlay, (iii) a compact \emph{fact pack} derived from the CNN, and (iv) the ECG Interpretation Guide as supporting knowledge. We provide the full guide for every ECG sample to reduce hallucinations and encourage guideline-consistent reasoning. The guide is injected as a fixed global context block; this approach differs from standard RAG (Vector DB retrieval) because the entire domain knowledge is condensed into a single document small enough to fit in the context window, eliminating the need for per-case retrieval.

\textbf{Fact pack.}
From $\mathbf{p}$ we construct a compact fact pack containing the full probabilities over the five superclasses and the top-$3$ predicted classes. This provides the LLM with an explicit representation of the CNN hypothesis set and confidence.

\textbf{Prompt design and evidence hierarchy.}
We enforce an evidence hierarchy in the system instruction: (1) Grad-CAM is the primary diagnostic evidence, (2) the ECG image contextualizes salient regions, (3) the fact pack is auxiliary, and (4) the guide supports guideline-consistent phrasing. The model is constrained to output exactly one JSON object.

\textbf{Output schema.}
The JSON contains: findings (morphology by lead/segment), impression (brief clinical summary),
evidence (lead-feature-rationale triples), consistency (aligned/partial/conflict), confidence, and recommendations. In evaluation, we use the impression field as the generated report sentence, while the remaining fields support qualitative auditing.

\section{Experiments}
\label{sec:experiments}

\subsection{Experimental setup}
 \subsubsection{Dataset}

We evaluate on the PTB-XL dataset~\cite{wagner2020ptbxl,strodthoff2021deeplearningptbxl}, which contains 21\,837 clinical 12-lead ECG records from 18\,885 patients. Each 10-second recording is annotated with one or more diagnostic statements, grouped into five diagnostic superclasses. Following prior work~\cite{strodthoff2021deeplearningptbxl,wu2024ecgllm}, we focus on: \emph{Normal}, \emph{Conduction Disturbance}, \emph{Hypertrophy}, \emph{Myocardial Infarction}, and \emph{ST/T Change}. We use the official stratified train/validation/test splits and represent targets as multi-hot vectors (multi-label setting).

\textbf{Reference text for report evaluation.}
For each ECG, we use the report field from PTB-XL database as the reference text. This field corresponds to the concatenation of the per-ECG diagnostic statements. Importantly, to avoid label leakage, we do not include any ground-truth diagnostic statements in the multimodal prompt during generation.

\subsubsection{Implementation details}

We include baseline for comparison.
\begin{enumerate}
  \item \textbf{Baseline~\cite{wu2024ecgllm} (CNN+Grad-CAM+MLLM)}: the multimodal LLM conditions on the ECG image, Grad-CAM overlay, and CNN fact pack, without the interpretation guide.
  \item \textbf{Ours (CNN+Grad-CAM+Guide+MLLM)}: The proposed method additionally provides the ECG Interpretation Guide as a fixed knowledge block for every ECG sample to improve guideline consistency and reduce hallucinations.
\end{enumerate}

We denoise waveforms using smoothing, a notch filter for power-line interference, and a high-pass filter for baseline wander removal, then render each ECG into a standardized 12-lead image. After that, We use ResNet-50 pretrained on ImageNet~\cite{deng2009imagenet,he2016deep} and fine-tune it for multi-label prediction ($K{=}5$ classes) with sigmoid outputs. We set the parameters by optimizing using Adam~\cite{kingma2014adam} with learning rate $10^{-3}$, batch size 128, for 50 epochs, selecting the best checkpoint using the validation split. After training the model, we generated Grad-CAM heatmaps~\cite{selvaraju2017gradcam} computed from the last convolutional layer for the top-$3$ predicted classes and overlayed on the ECG image with transparency $\alpha=0.45$. We constructed the ECG Interpretation Guide by summarizing knowledge in medical books \cite{hampton2019ecg,sajjan2013learn}, using an offline lightweight LLM as described in Stage~2. 

We invoke a Gemini-based multimodal model (gemini-2.5-flash-lite)~\cite{geminiteam2023gemini} via the vendor API endpoint with temperature $0.2$, a maximum of 1200 output tokens, and top-3 sampling ($top\_k=3$) where applicable. For each ECG, the API request includes the Grad-CAM overlay (scaled to standard high-resolution), the processed ECG image, and a fact pack containing the full CNN probabilities and the top-$3$ predicted classes. In the proposed method, the guide is appended as an additional knowledge block for every sample; in the baseline, this block is removed. We generate reports over the official test split and analyze an additional 200-sample qualitative subset. Requests are batched for efficiency, and we discard outputs that do not satisfy the required JSON schema.

\subsubsection{Evaluation metrics}

We compute BERTScore~\cite{zhang2020bertscore} (Precision/Recall/F1) between generated impression strings and the reference diagnostic statement strings from the raw data. Since references are in German while our generated impressions are in English, we report BERTScore using two complementary settings:
\begin{enumerate}
  \item \textbf{Cross-lingual BERTScore:} directly score German references vs.\ English generations.
  \item \textbf{Translated-reference BERTScore:} first translate each German reference into English using Google Translate  , then compute BERTScore between the translated English references and English generations.
\end{enumerate}
Our \emph{main reported BERTScore} is the arithmetic mean of the two settings above (reported per metric and then averaged over the test set). We additionally report each setting separately for transparency. We compute BERTScore using the \texttt{xlm-roberta-large} model (layer 24) without IDF weighting or baseline rescaling, which supports both German and English contexts robustly.

On the qualitative subset, we use a forced-choice judging protocol~\cite{zheng2024judging}: a blinded LLM judge is given the ECG context, the reference statement, and two candidate reports (baseline vs.\ ours), and must choose exactly one winner (no ties). The judge is blinded to method identity, and report order is randomized. Specifically, we employ per-sample A/B randomization with a hidden mapping to ensure true blinding. Additionally, the judge prompt and decoding settings (temperature=0.0) remain fixed to prevent inference artifacts. To reduce variance, we repeat the entire judging procedure 5 times with different randomized report orderings and report the mean win-rate across runs (and the standard deviation when applicable). We summarize results using win-rate statistics (mean over 5 runs).

\subsection{Experiment Results}
\label{sec:results}

\subsubsection{CNN Backbone selection on 5k samples}

\begin{table}[t]
  \centering
  \caption{Backbone comparison trained on training set}
  \label{tab:backbone_5k}
  \begin{tabular}{lccc}
    \toprule
    Model & Precision & Recall & F1-score \\
    \midrule
    DenseNet~\cite{huang2017densenet} & 0.71 & 0.66 & 0.68 \\
    Inception-v3~\cite{szegedy2016rethinking} & 0.56 & 0.35 & 0.43 \\
    ResNet-50~\cite{he2016deep} & \textbf{0.85} & \textbf{0.79} & \textbf{0.82} \\
    VGG-16~\cite{simonyan2015very} & 0.64 & 0.43 & 0.49 \\
    \bottomrule
  \end{tabular}
\end{table}

Table~\ref{tab:backbone_5k} shows that ResNet-50 achieves the best overall performance on the 5k-sample subset (F1 = 0.82). We therefore select ResNet-50 as the CNN backbone for the full pipeline.

\begin{table}[t]
  \centering
  \caption{Per-class performance of ResNet-50 }
  \label{tab:perclass_resnet_5k}
  \begin{tabular}{lcccc}
    \toprule
    Class & Precision & Recall & F1-score & Instances \\
    \midrule
    Norm & 0.89 & 0.71 & 0.79 & 239 \\
    MI   & 0.95 & 0.44 & 0.60 & 136 \\
    STTC & 0.72 & 0.88 & 0.79 & 244 \\
    CD   & 0.88 & 0.92 & 0.90 & 587 \\
    HYP  & 0.91 & 0.72 & 0.80 & 276 \\
    \bottomrule
  \end{tabular}
\end{table}

Table~\ref{tab:perclass_resnet_5k} indicates strong performance for CD (F1 = 0.90) and stable results on Norm/STTC/HYP. MI shows lower recall (0.44), suggesting infarction patterns are harder to capture under limited data or affected by class imbalance. This motivates grounding the final text explanation in guideline-style knowledge to improve trustworthiness.

\subsubsection{Report Quality (BERTScore)}

We evaluate reports using BERTScore \cite{zhang2020bertscore} against the reference diagnostic statement strings (the report field in PTB-XL). BERTScore is computed on the official PTB-XL test split and averaged over all samples with valid generated outputs (valid JSON and a non-empty impression field). Because references are in German while generated impressions are in English, we compute BERTScore under two complementary settings (cross-lingual German$\leftrightarrow$English and translated-reference German$\rightarrow$English). We report the setting-specific results separately below.

\begin{table}[b]
  \centering
  \caption{Cross-lingual BERTScore on the PTB-XL test split (German reference vs.\ English generation)}
  \label{tab:bertscore_crosslingual}
  \begin{tabular}{lccc}
    \toprule
    Method & Precision & Recall & F1 \\
    \midrule
    Baseline~\cite{wu2024ecgllm} & 0.816 & 0.821 & 0.818 \\
    Ours     & \textbf{0.984} & \textbf{0.981} & \textbf{0.982} \\
    \bottomrule
  \end{tabular}
\end{table}

\begin{table}
  \centering
  \caption{LLM-based Automated pairwise preference on the qualitative subset }
  \label{tab:llm_referee}
  \begin{tabular}{lcc}
    \toprule
    Method & Gemini win-rate  &  GPT-4o-mini win-rate  \\
    \midrule
    Baseline~\cite{wu2024ecgllm} & 38\%  & 24\%  \\
    Ours     & \textbf{62\%} & \textbf{76\%} \\
    \bottomrule
  \end{tabular}
\end{table}

\begin{figure}[t]
  \centering
  \includegraphics[width=\textwidth]{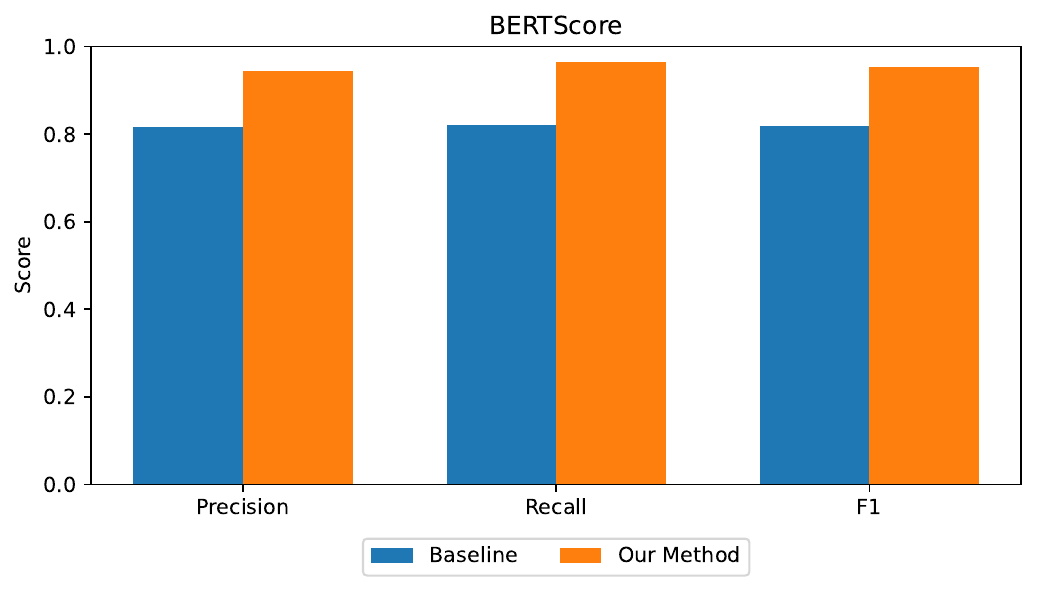}
  \caption{BERTScore comparison between the baseline and our guide-grounded method.}
  \label{fig:bertscore}
\end{figure}

\begin{table}[t]
  \centering
  \caption{Translated-reference BERTScore on the PTB-XL test split (German$\rightarrow$English translated reference vs.\ English generation)}
  \label{tab:bertscore_translated}
  \begin{tabular}{lccc}
    \toprule
    Method & Precision & Recall & F1 \\
    \midrule
    Baseline~\cite{wu2024ecgllm} & 0.816 & 0.821 & 0.818 \\
    Ours     & \textbf{0.944} & \textbf{0.964} & \textbf{0.953} \\
    \bottomrule
  \end{tabular}
\end{table}
As a sanity check, we randomly permute hypothesis--reference pairs; the BERTScore drops sharply, confirming the metric is not inflated by artifacts.

\subsubsection{LLM-based Automated Evaluation }
To complement embedding-based metrics, we adopt an LLM-based Automated protocol~\cite{zheng2024judging} on a qualitative subset. We use forced-choice pairwise comparison between the baseline report and our report . Among 200 sampled ECGs, 192 produce valid judge outputs and are included in the final statistics. To reduce variance, we repeat the entire judging procedure 5 times with different randomized report orderings and report the mean win-rate across runs (and the standard deviation when applicable) (mean over 5 runs).

Using a Gemini-based judge~\cite{geminiteam2023gemini}, our method is preferred in 62.0\% of cases (mean over 5 runs), while the baseline is preferred in 38.0\%. Using a GPT-4o-mini judge, our method is preferred in 76.0\% of cases (mean over 5 runs), while the baseline is preferred in 24.0\%. These results support that injecting guideline-style knowledge improves perceived correctness and usefulness of the generated reports.

\subsubsection{Qualitative Analysis}

Qualitative inspection suggests that baseline reports tend to be brief and occasionally generic, while the guide-grounded method more consistently uses guideline-style criteria and produces clearer lead- and feature-specific rationale aligned with Grad-CAM salient regions. In myocardial infarction cases, the baseline often mentions ST changes without sufficient diagnostic context, whereas the guide-grounded reports more frequently include criterion-like descriptions. In rhythm-related cases, the guide grounding reduces unsupported claims and improves consistency between the final impression and the highlighted waveform evidence. A small expert review of randomly sampled reports further supports improved factual consistency and clinician confidence.

\FloatBarrier
\subsection{Discussion}

Our experiments show that grounding multimodal LLMs in ECG textbooks\cite{hampton2019ecg,sajjan2013learn} and clinical guidelines via \emph{guide injection} improves the semantic quality and perceived factual consistency of generated reports without sacrificing classification accuracy. The gains are substantial across BERTScore components and are corroborated by LLM-based Automated scores. A key strength of our approach is modularity: the CNN classifier, Grad-CAM explainer, guide distillation procedure, and MLLM can each be improved independently. For example, a better CNN architecture or a domain-specific ECG-language model~\cite{zhao2024ecgchat,liu2024teachmlmecg} could be plugged into the pipeline without redesigning the overall framework.

However, several limitations remain. Firstly, since our guides depend on the quality of the injected medical books\cite{hampton2019ecg,sajjan2013learn}, and over time, they will become outdated, so we need to constantly update the guide. Secondly, injecting the full guide increases the input token count compared to minimal prompts. However, the guide is a fixed context block (no per-case retrieval), which keeps the inference pipeline simple and reproducible. Finally, LLM-based Automated evaluation, while helpful, is still an approximation to human expert assessment and may inherit biases from the underlying model.

\vspace{-0.1in}
\FloatBarrier
\section{Conclusions}

We presented a guide-grounded multimodal framework for explainable ECG diagnosis that combines CNN-based classification, Grad-CAM visual explanations, and guideline-grounded report generation via multimodal LLMs. A key component is an ECG Interpretation Guide synthesized by curated textbooks\cite{hampton2019ecg,sajjan2013learn} and guidelines, which is then injected into the prompt for every sample to reduce hallucinations and encourage guideline-consistent reporting. By conditioning the LLM on this guide, our method significantly improves the semantic quality and perceived consistency of generated reports while preserving strong classification performance on PTB-XL. Additional evaluation with LLM-based Automated further supports the superiority of our guide-grounded reports over a strong CNN+Grad-CAM+MLLM baseline. This work illustrates a practical path towards trustworthy, explainable AI-assisted cardiac diagnosis. Despite these improvements over previous approaches, further work is required to enhance the efficiency and scalability of the proposed framework.
\vspace{-0.1in}

\section*{Acknowledgement}
This research is supported by the Ben Dam Me Award Fund, the Vietnam Young Talent Support Fund, and the Number One Brand, Tan Hiep Phat Group.

\bibliographystyle{splncs04}
\bibliography{refs}

@article{wagner2020ptbxl,
  title   = {{PTB-XL}: A large publicly available electrocardiography dataset},
  author  = {Wagner, Patrick and Strodthoff, Nils and Bousseljot, Ralf-Dieter and Kreiseler, Dieter and Lunze, Fatima I and Samek, Wojciech and Schaeffter, Tobias},
  journal = {Scientific Data},
  volume  = {7},
  number  = {1},
  pages   = {154},
  year    = {2020},
  
}

@inproceedings{deng2009imagenet,
  title     = {{ImageNet}: A large-scale hierarchical image database},
  author    = {Deng, Jia and Dong, Wei and Socher, Richard and Li, Li-Jia and Li, Kai and Fei-Fei, Li},
  booktitle = {Proceedings of the IEEE Conference on Computer Vision and Pattern Recognition (CVPR)},
  pages     = {248--255},
  year      = {2009},
 

}

@inproceedings{he2016deep,
  title     = {Deep residual learning for image recognition},
  author    = {He, Kaiming and Zhang, Xiangyu and Ren, Shaoqing and Sun, Jian},
  booktitle = {Proceedings of the IEEE Conference on Computer Vision and Pattern Recognition (CVPR)},
  pages     = {770--778},
  year      = {2016},
 
}

@inproceedings{huang2017densenet,
  title     = {Densely connected convolutional networks},
  author    = {Huang, Gao and Liu, Zhuang and van der Maaten, Laurens and Weinberger, Kilian Q},
  booktitle = {Proceedings of the IEEE Conference on Computer Vision and Pattern Recognition (CVPR)},
  pages     = {4700--4708},
  year      = {2017},
 
}

@inproceedings{szegedy2016rethinking,
  title     = {Rethinking the Inception Architecture for Computer Vision},
  author    = {Szegedy, Christian and Vanhoucke, Vincent and Ioffe, Sergey and Shlens, Jonathon and Wojna, Zbigniew},
  booktitle = {Proceedings of the IEEE Conference on Computer Vision and Pattern Recognition (CVPR)},
  pages     = {2818--2826},
  year      = {2016},

}

@inproceedings{simonyan2015very,
  title     = {Very Deep Convolutional Networks for Large-Scale Image Recognition},
  author    = {Simonyan, Karen and Zisserman, Andrew},
  booktitle = {International Conference on Learning Representations (ICLR)},
  year      = {2015},

}

@article{openai2023gpt4,
  title   = {{GPT-4} Technical Report},
  author  = {{OpenAI}},
  journal = {arXiv preprint arXiv:2303.08774},
  year    = {2023},
  }

@article{geminiteam2023gemini,
  title   = {Gemini: A Family of Highly Capable Multimodal Models},
  author  = {{Gemini Team} and Anil, Rohan and Borgeaud, Sebastian and Wu, Yonghui and others},
  journal = {arXiv preprint arXiv:2312.11805},
  year    = {2023},
  }

@inproceedings{wu2024ecgllm,
  title     = {Enhancing Explainability of Deep Learning-Based {ECG} Diagnosis Using Large Language Models},
  author    = {Wu, Shi and Zhou, Jianlong and Dong, Yifei and Chen, Fang},
  booktitle = {Proceedings of the 8th International Conference on Advances in Artificial Intelligence (ICAAI~'24)},
  pages     = {61--65},
  year      = {2024},
  publisher = {ACM},
 
}

@article{zhao2024ecgchat,
  title   = {{ECG-Chat}: A Large {ECG}-Language Model for Cardiac Disease Diagnosis},
  author  = {Zhao, Yubao and Zhang, Tian and Wang, Xu and Han, Puyu and Chen, Tong and Huang, Linlin and Jin, Youzhu and Kang, Jiaju},
  journal = {arXiv preprint arXiv:2408.08849},
  year    = {2024}
}

@article{liu2024teachmlmecg,
  title   = {Teach Multimodal {LLMs} to Comprehend Electrocardiographic Images},
  author  = {Liu, Ruoqi and Bai, Yuelin and Yue, Xiang and Zhang, Ping},
  journal = {arXiv preprint arXiv:2410.19008},
  year    = {2024}
}

@inproceedings{selvaraju2017gradcam,
  title     = {Grad-{CAM}: Visual Explanations from Deep Networks via Gradient-based Localization},
  author    = {Selvaraju, Ramprasaath R and Cogswell, Michael and Das, Abhishek and Vedantam, Ramakrishna and Parikh, Devi and Batra, Dhruv},
  booktitle = {Proceedings of the IEEE International Conference on Computer Vision (ICCV)},
  pages     = {618--626},
  year      = {2017},
 }

@article{strodthoff2021deeplearningptbxl,
  title   = {Deep Learning for {ECG} Analysis: Benchmarks and Insights from {PTB}-{XL}},
  author  = {Strodthoff, Nils and Wagner, Patrick and Schaeffter, Tobias and Samek, Wojciech},
  journal = {IEEE Journal of Biomedical and Health Informatics},
  volume  = {25},
  number  = {5},
  pages   = {1519--1528},
  year    = {2021},
  }

@inproceedings{ribeiro2016lime,
  title     = {{``Why Should I Trust You?''}: Explaining the Predictions of Any Classifier},
  author    = {Ribeiro, Marco Tulio and Singh, Sameer and Guestrin, Carlos},
  booktitle = {Proceedings of the 22nd ACM SIGKDD International Conference on Knowledge Discovery and Data Mining},
  pages     = {1135--1144},
  year      = {2016}
}

@inproceedings{lundberg2017shap,
  title     = {A Unified Approach to Interpreting Model Predictions},
  author    = {Lundberg, Scott M and Lee, Su-In},
  booktitle = {Advances in Neural Information Processing Systems},
  volume    = {30},
  year      = {2017}
}

@inproceedings{zhang2020bertscore,
  title     = {{BERTScore}: Evaluating Text Generation with {BERT}},
  author    = {Zhang, Tianyi and Kishore, Varsha and Wu, Felix and Weinberger, Kilian Q and Artzi, Yoav},
  booktitle = {International Conference on Learning Representations (ICLR)},
  year      = {2020}
}

@article{rajpurkar2017cardiologist,
  title   = {Cardiologist-Level Arrhythmia Detection with Convolutional Neural Networks},
  author  = {Rajpurkar, Pranav and Hannun, Awni Y and Haghpanahi, Masoumeh and Bourn, Codie and Ng, Andrew Y},
  journal = {arXiv preprint arXiv:1707.01836},
  year    = {2017}
}

@article{hicks2021ecggradcam,
  title   = {Explaining Deep Neural Networks for Knowledge Discovery in Electrocardiogram Analysis},
  author  = {Hicks, Steven A and Isaksen, Jonas L and Thambawita, Vajira and Ghouse, Jonas and Ahlberg, Gustav and Linneberg, Allan and others},
  journal = {Scientific Reports},
  volume  = {11},
  number  = {1},
  pages   = {10949},
  year    = {2021},
  }

@article{jahmunah2022mi_gradcam,
  title   = {Explainable Detection of Myocardial Infarction Using Deep Learning Models with Grad-{CAM} Technique on {ECG} Signals},
  author  = {Jahmunah, Vicneswary and Ng, Eddie YK and Tan, Ru-San and Oh, Shu Lih and Acharya, U Rajendra},
  journal = {Computers in Biology and Medicine},
  volume  = {146},
  pages   = {105550},
  year    = {2022},
  }

@article{kingma2014adam,
  title   = {Adam: A Method for Stochastic Optimization},
  author  = {Kingma, Diederik P and Ba, Jimmy},
  journal = {arXiv preprint arXiv:1412.6980},
  year    = {2014},

}

@article{butterworth1930theory,
  title   = {On the Theory of Filter Amplifiers},
  author  = {Butterworth, Stephen},
  journal = {Wireless Engineer},
  volume  = {7},
  number  = {6},
  pages   = {536--541},
  year    = {1930},

}

@article{zheng2024judging,
  title   = {Judging LLM-as-a-Judge with MT-Bench and Chatbot Arena},
  author  = {Zheng, Lianmin and Chiang, Wei-Lin and Sheng, Ying and Zhuang, Soheil and Wu, Junxian and Zhuang, Wei and others},
  journal = {arXiv preprint arXiv:2306.05685},
  year    = {2023}
}

@book{hampton2019ecg,
  title     = {The ECG Made Easy},
  author    = {Hampton, John R. and Hampton, Joanna},
  edition   = {9th},
  year      = {2019},
  publisher = {Elsevier}
}

@book{sajjan2013learn,
  title     = {Learn ECG in a Day: A Systematic Approach},
  author    = {Sajjan, M.},
  year      = {2013},
  publisher = {Jaypee Brothers Medical Publishers}
}

\end{document}